%% file: main.tex
\documentclass[letterpaper]{article} 
\usepackage[draft]{aaai25}  
\usepackage{times}  
\usepackage{helvet}  
\usepackage{courier}  
\usepackage[hyphens]{url}  
\usepackage{graphicx} 
\usepackage{natbib}  
\usepackage{caption} 
\usepackage{algorithm}
\usepackage{algorithmic}

\usepackage{newfloat}
\usepackage{listings}
\DeclareCaptionStyle{ruled}{labelfont=normalfont,labelsep=colon,strut=off} 
\floatstyle{ruled}
\newfloat{listing}{tb}{lst}{}
\floatname{listing}{Listing}
\title{Intelligent OPC Engineer Assistant for Semiconductor Manufacturing}
\author{
    Guojin Chen\textsuperscript{\rm 1,2}\thanks{Work accomplished during internship at NVIDIA.},
    Haoyu Yang\textsuperscript{\rm 2},
    Bei Yu\textsuperscript{\rm 1}
    Haoxing Ren\textsuperscript{\rm 2},
}
\affiliations{
    \textsuperscript{\rm 1}Chinese University of Hong Kong\\
    \textsuperscript{\rm 2}NVIDIA\\
}

\usepackage{bibentry}

\usepackage{xcolor,colortbl}
\usepackage{amssymb}
\usepackage{amsmath}
\usepackage{cleveref}
\usepackage{subfig}
\usepackage{booktabs, multirow} 
\usepackage{soul}
\usepackage{changepage,threeparttable} 
\usepackage[most]{tcolorbox}
\graphicspath{{./figs/}}

\usepackage{tikz}
\newcommand*\circled[1]{\tikz[baseline=(char.base)]{ \node[shape=circle,draw,inner sep=0.2pt] (char) {#1};}}

\begin{document}

\maketitle

\input{doc/abstract}
\input{doc/intro}

\input{doc/prelim}
\input{doc/algo}

\input{doc/exp}
\input{doc/conclu}

{
	\bibliography{refs/sim.bib,refs/llm.bib,refs/recipe.bib,refs/dfm.bib}
}

\input{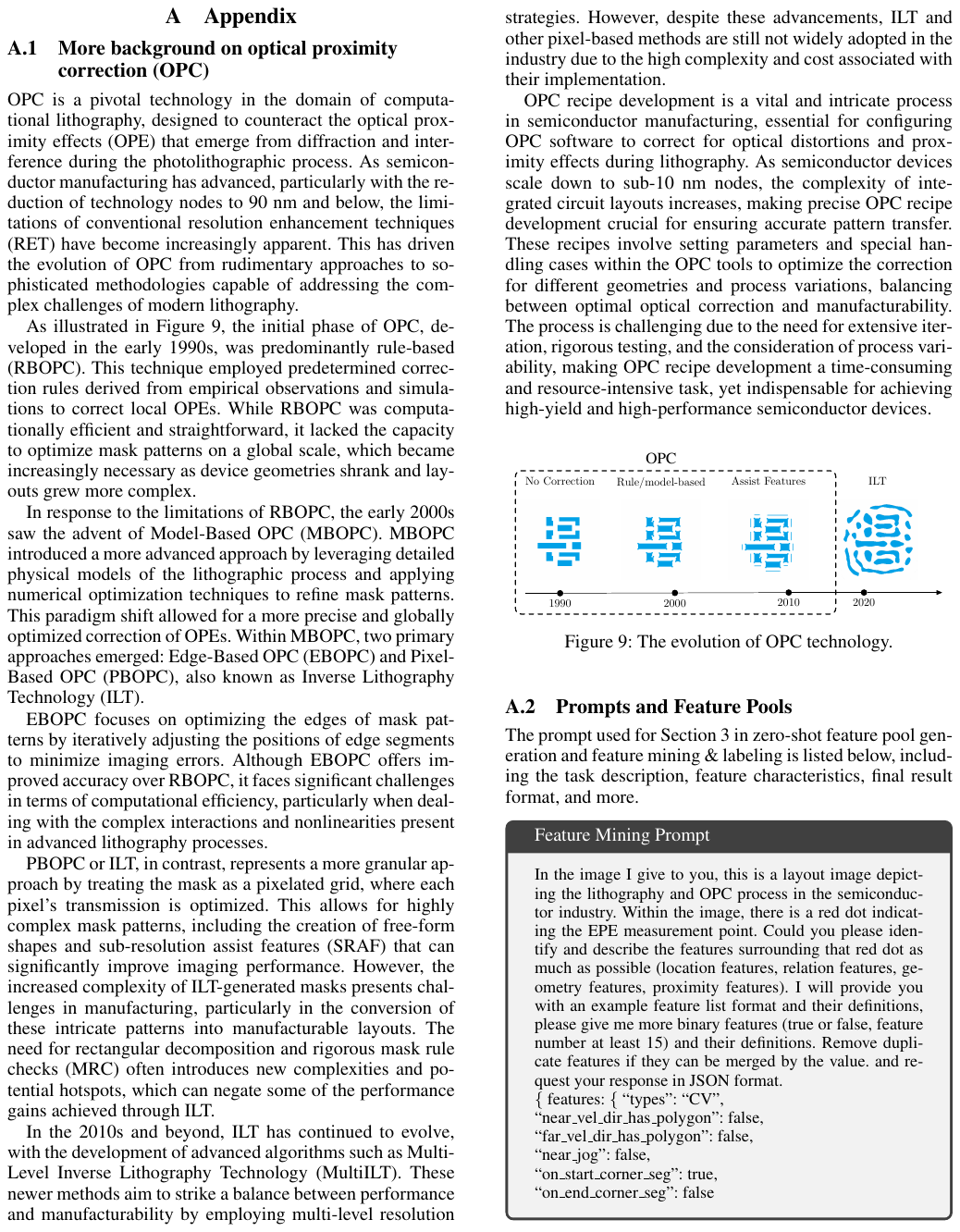}
\end{document}

%% file: doc/abstract.tex
\begin{abstract}
Advancements in chip design and manufacturing have enabled the processing of complex tasks such as deep learning and natural language processing, paving the way for the development of artificial general intelligence (AGI). AI, on the other hand, can be leveraged to innovate and streamline semiconductor technology from planning and implementation to manufacturing. In this paper, we present \textit{Intelligent OPC Engineer Assistant}, an AI/LLM-powered methodology designed to solve the core manufacturing-aware optimization problem known as optical proximity correction (OPC). The methodology involves a reinforcement learning-based OPC recipe search and a customized multi-modal agent system for recipe summarization. Experiments demonstrate that our methodology can efficiently build OPC recipes on various chip designs with specially handled design topologies, a task that typically requires the full-time effort of OPC engineers with years of experience.
\end{abstract}

%% file: doc/intro.tex
\section{Introduction}
\label{sec:intro}
Recent advancements in chip design and manufacturing have enabled the handling of intricate tasks such as deep learning and natural language processing, bringing us closer to achieving artificial general intelligence (AGI).
Meanwhile, AI is revolutionizing semiconductor technology by optimizing every stage of the semiconductor lifecycle, from conceptual planning to execution and manufacturing.
By leveraging AI-driven innovations, the industry can achieve greater efficiency, precision, and performance, potentially accelerating the development of next-generation chips.

\begin{figure}[tb!]
  \centering
  \includegraphics[width=.96\linewidth]{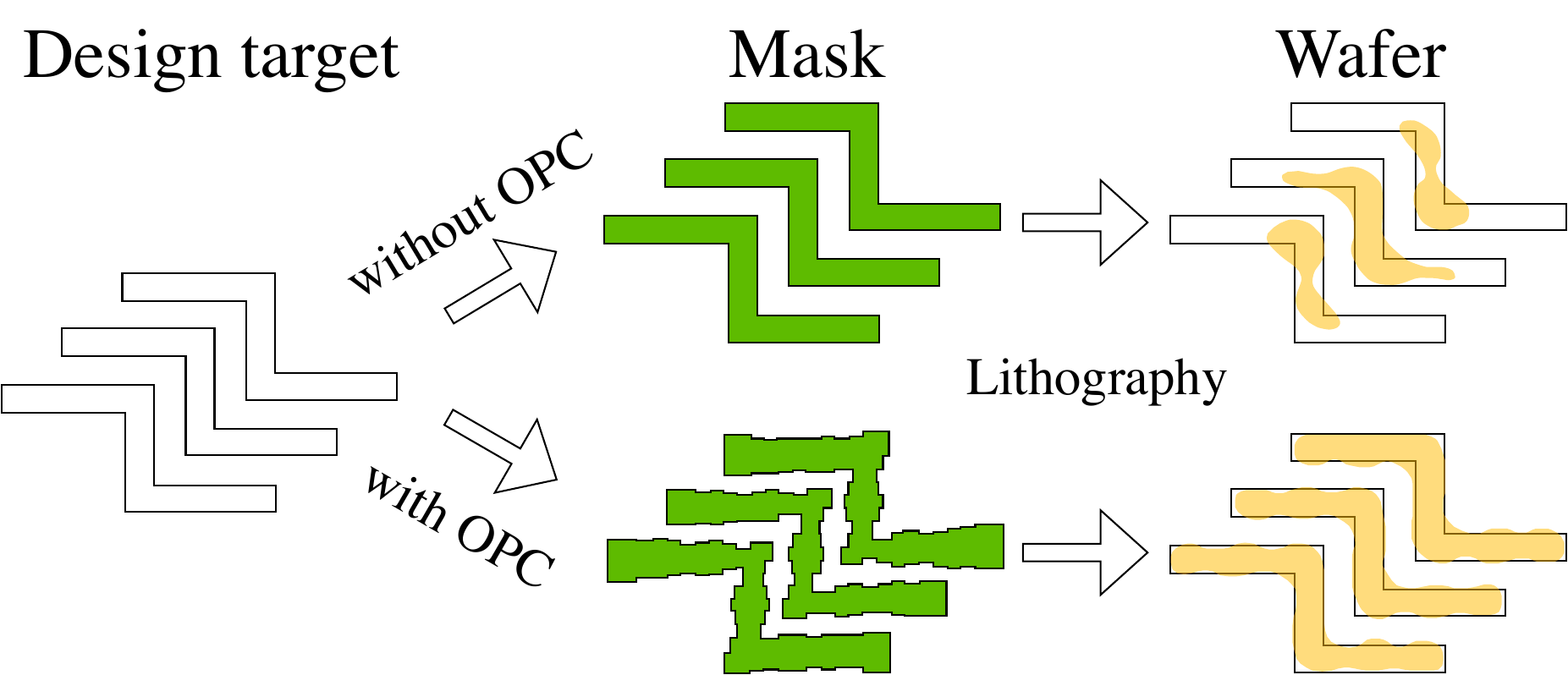}
  \caption{Motivation for optical proximity correction (OPC) in semiconductor manufacturing process.}
  \label{fig:opc-motiv}
\end{figure}

In this paper, we utilize AI to address the complexities of optical proximity correction (OPC), a critical process in semiconductor manufacturing.
As shown in \Cref{fig:opc-motiv}, OPC involves adjusting chip designs to counteract lithography distortions, ensuring that the final patterns on the silicon wafer closely match the intended design with high precision.
As depicted in \Cref{fig:recipe-engine}, a complete OPC solution comprises both the solver and the recipe. The OPC solver includes the core OPC algorithms, such as lithography imaging computation, mask database management, gradient calculation, and shape perturbation, among other processes~\cite{MEEF-Photomask2002-Granik,MEEF-lei2014model}.
Foundries and EDA (Electronic Design Automation) vendors each have their own OPC solver implementations, which are built on advanced algorithms and follow similar workflows~\cite{TOOL-calibre-OPC,TOOL-Proteus-OPC}.
The OPC recipe, however, contains specific configurations tailored to optimize a particular design. It includes common optimization parameters such as step size, maximum iterations, shape movement constraints, polygon fragmentation policies, and error control strategies. Recipes also incorporate specialized rules for handling unique chip design patterns that cannot be effectively addressed through standard optimization settings. These specialized rules are crucial in OPC and are typically developed by experienced OPC engineers through extensive trial and error.

To enhance the efficiency of chip development, we present the \textit{Intelligent OPC Engineer Assistant}, an AI-driven framework designed to assist human engineers in the rapid development of OPC recipes.
This methodology integrates a reinforcement learning (RL)-based approach for optimizing objective searches and a multi-modality large language model (MLLM)-backboned agent system to facilitate spatial reasoning and recipe summarization.

The remainder of the manuscript is organized as follows.
In \Cref{sec:prelim}, we delve into the background of OPC and recipe development, highlighting several works that address related challenges using RL and LLM agents.
In \Cref{sec:method}, we present the core framework of our algorithm, structured in a two-stage approach.
Experimental results and conclusions are provided in \Cref{sec:exp} and \Cref{sec:conclusion}, respectively.
Additional prompts, algorithm details, and comprehensive explanations are included in the appendix.

%% file: doc/prelim.tex
\section{Preliminaries}
\label{sec:prelim}
\subsection{Related Works}
\label{subsec:related_work}

\begin{figure}[tb!]
  \centering
  \includegraphics[width=0.97\linewidth]{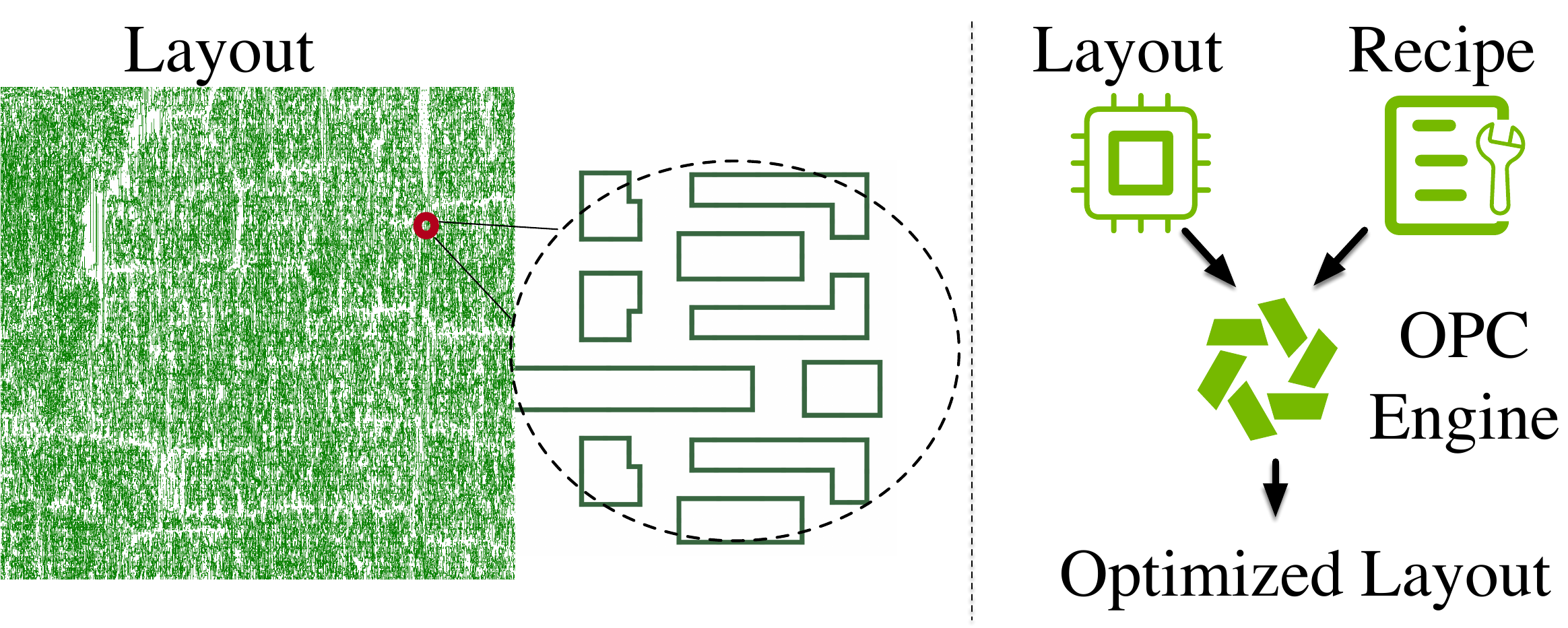}
  \caption{Left: The full-chip layout. Right: The relationship between the layout, OPC Recipe and OPC Engine.}
  \label{fig:recipe-engine}
\end{figure}

\subsubsection{OPC and Reinforcement Learning}
As device geometries shrink to the nanometer scale, the limitations of traditional optical lithography become apparent, necessitating advanced computational methods to achieve the desired fidelity.
As illustrated in \Cref{fig:opc-motiv}, OPC is a critical component of computational lithography, addressing distortions and proximity effects that arise during the lithographic process.
By systematically adjusting the mask design to counteract these effects, OPC ensures that the final printed patterns on the wafer closely match the intended design.
This correction is pivotal for maintaining device performance, yield, and reliability in the semiconductor industry.
The integration of computational lithography and OPC is thus crucial for advancing semiconductor technology, enabling the production of increasingly smaller and more complex integrated circuits.
In recent years, numerous studies have utilized AI and ML to enhance OPC algorithms~\cite{pmlr-v235-yang24s,L2OILT_TCAD,OPC-ICCAD2020-DAMO}.
Notably, \cite{liang2023rl,liang2024camo} have proposed reinforcement learning-based OPC approaches that integrate spatial correlations and OPC-specific principles, enhancing performance and efficiency across both metal and via layers.

\subsubsection{OPC Recipe Development and LLM Agents}
\begin{figure}[!tbp]
  \centering
  \includegraphics[width=0.86\linewidth]{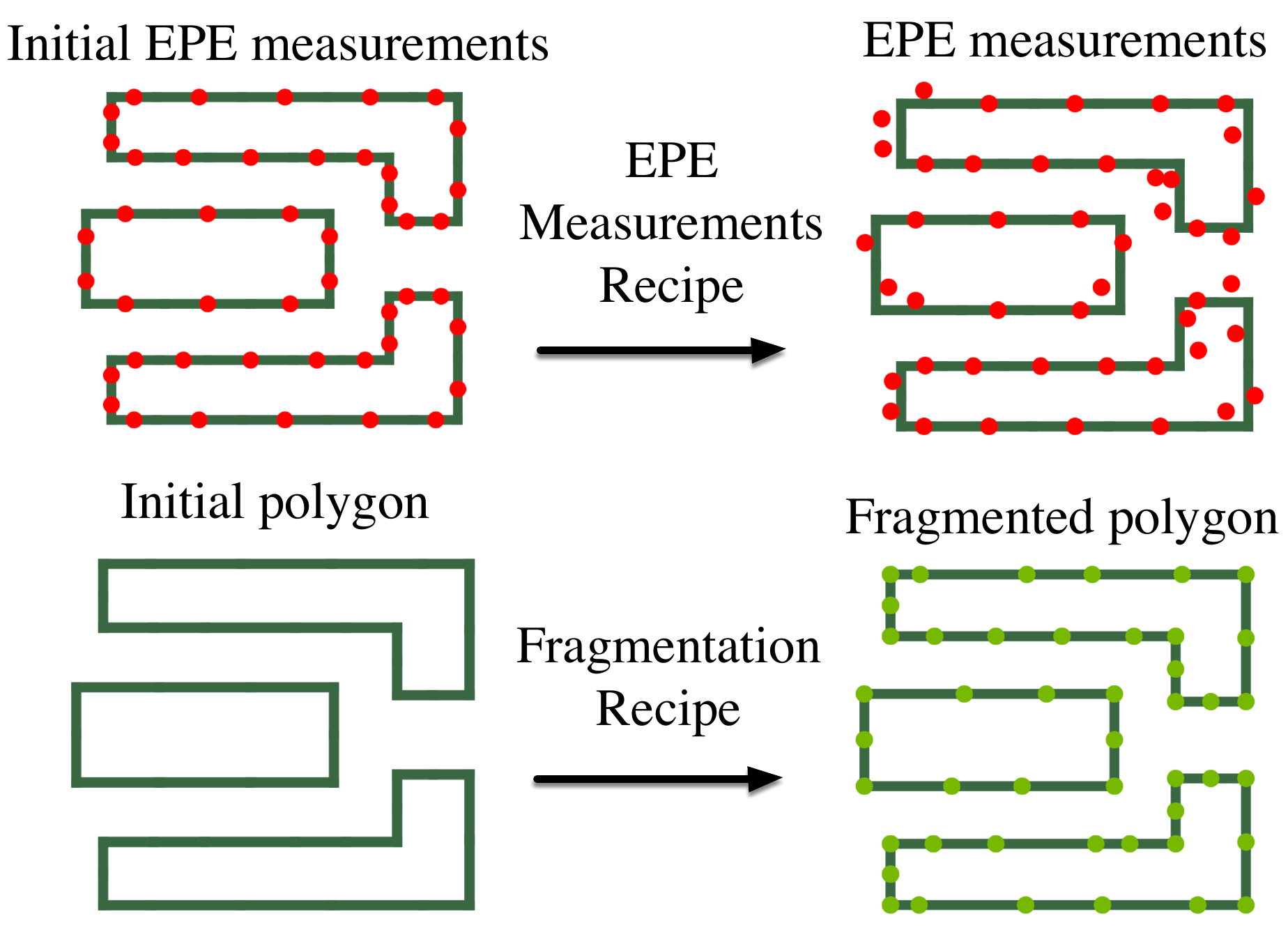}
  \caption{
      OPC recipe development.
      The upper figure shows the recipe for optimizing EPE measurement points, while the lower figure shows the edge fragmentation recipe.
  }
  \label{fig:opc-recipe}
\end{figure}

As illustrated in \Cref{fig:recipe-engine}, the primary motivation for OPC recipe development is to enhance the productivity of OPC engineers in the context of modern semiconductor manufacturing.
As depicted in \Cref{fig:opc-recipe}, once the OPC engine is well-established, it becomes necessary to adjust the recipe based on various factors such as pattern characteristics, location, and process nodes, in addition to optimizing the algorithm's intrinsic parameters.
\Cref{fig:opc-recipe} illustrates two pertinent examples from this study.
The first example concerns the adjustment of EPE measurement points, depicted as red dots in \Cref{fig:opc-recipe}.
Initially, these measurement points are distributed along the boundaries, leading to excessive optimization at the corners of the patterns.
This over-optimization can degrade the overall effectiveness of the OPC process.
Therefore, an optimized EPE measurement recipe involves relocating the measurement points either outward or inward from the corners to prevent excessive optimization at specific locations.
Traditionally, this adjustment relied heavily on the engineers' experience and experimental tuning,
which was time-consuming and costly, especially with the advent of new technology nodes.
The second example involves the fragmentation of polygons, which is essential for the OPC engine to function.
The recipe plays a crucial role in fragmenting different polygons according to their characteristics.
The illustration in \Cref{fig:opc-recipe} shows the initial polygon and its fragmented version post-recipe application,
demonstrating how the fragmentation recipe is tailored to ensure optimal performance of the OPC engine.
As design complexity and the number of device and mask layers increase, traditional OPC methods have become inadequate for advanced nodes,
necessitating the creation of more customized OPC recipes.
This growing complexity has led to a surge in the engineering workload,
demanding more OPC engineers to handle the sophisticated new OPC recipes~\cite{BBS-recipe-Wu-2016,asthana2016opc,qingwei-opc-cost-2010}.
Recently, The integration of LLMs as agents to automate the EDA process has recently attracted considerable research interest. For instance, RTLFixer~\cite{tsai2023rtlfixer} focuses on automated debugging and code repair, while ChipNeMo~\cite{liu2023chipnemo} serves as an engineering assistant chatbot, facilitating EDA script generation and bug analysis. Additionally, ChatEDA~\cite{wu2024chateda} incorporates LLMs into traditional design workflows, allowing designers to interact with EDA tools through a conversational interface using natural language. These advancements significantly enhance the automation of EDA processes and improve engineering efficiency.

\subsection{Evaluation Metrics for OPC}
\label{subsec:opc-metrics}
\begin{figure}[!tb]
  \centering
  \includegraphics[width=0.98\linewidth]{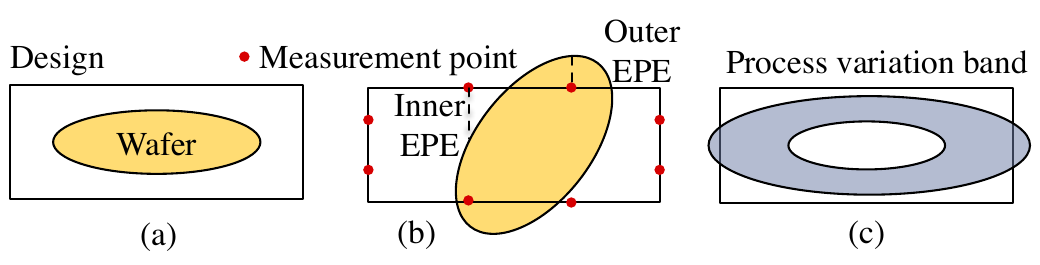}
  \caption{Evaluation metrics for OPC.
  (a) Definition of design target and the wafer pattern.
  (b) Illustration of EPE measurement points, including inner and outer EPE violations for calculating the total EPE count and EPE distance.
  (c) Process variation band (PVB) calculation.}
  \label{fig:opc-metrics}
\end{figure}

In this paper, we use domain-specific process variation band (PVB) and edge placement error (EPE) as two typical metrics to evaluate OPC performance.
As shown in \Cref{fig:opc-metrics}, these metrics provide a comprehensive assessment of the quality of the OPC mask,
capturing different aspects of the lithographic process.
\begin{description}
    \item[Edge placement error (EPE)] 
        quantifies the geometric distortion of the resist image~\cite{OPC-ICCAD2013-Banerjee}. It is calculated by sampling points along the edges of target shapes and counting the number of points where the distance between the target and printed pattern exceeds a predefined threshold.
  The smaller the EPE, the better the OPC mask quality.
  In this paper, we calculate the EPE number and EPE distance to evaluate EPE.
  The EPE number (EPE N) is counted by the number of points where the distance between the target and printed pattern exceeds 1 $nm$.
  The EPE distance (EPE D) is summed by the distance between the target and printed pattern where the distance exceeds 1 $nm$.
  \item[PVB] evaluates the robustness of the mask against different process conditions~\cite{OPC-ICCAD2013-Banerjee}.
  PVB represents the range of deviations that can occur during the patterning of semiconductor features.
  As illustrated in \Cref{fig:opc-metrics}(c), this band defines the tolerances within which the process must operate to achieve the desired feature dimensions and ensure product quality. A narrower process variation band indicates better process control, crucial for manufacturing advanced semiconductor devices.
  It is computed by measuring the bitwise XOR region between the printed images from the maximum and minimum process conditions ($\pm 2\%$) $\boldsymbol{Z}_{max}, \boldsymbol{Z}_{min}$.
  $\text{PVB}(\boldsymbol{Z}_{max}, \boldsymbol{Z}_{min}) = \Vert \boldsymbol{Z}_{max} - \boldsymbol{Z}_{min} \Vert_2^2.   \label{eqn:pvb}$
\end{description}

\subsection{Problem Formulation}
\label{subsec:problem}
By automating these processes, this paper seeks to reduce the time and cost involved in OPC recipe development, thereby improving the overall efficiency and effectiveness of computational lithography in semiconductor manufacturing.
The study has two primary objectives: first, to develop an automated method for generating a comprehensive set of recipes applicable to full-chip patterns, including the optimization of EPE measurement points and layout fragmentation points;
second, to ensure that these recipes minimize both the PVB and the EPE.

%% file: doc/algo.tex
\section{Methodology}
\label{sec:method}

\subsection{Overview: A Two-stage Approach}
\begin{figure*}[tb!]
    \centering
    \includegraphics[width=0.88\textwidth]{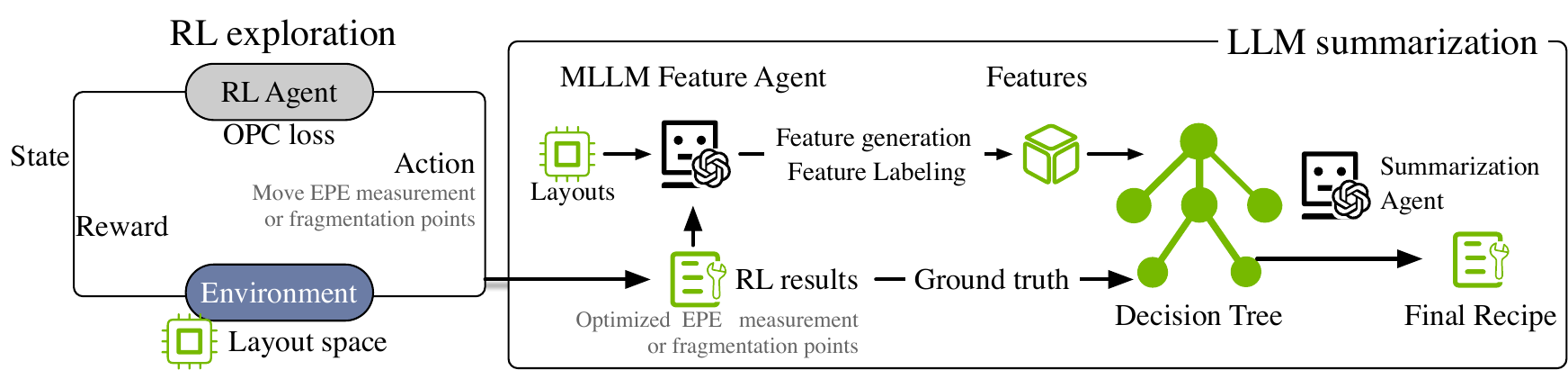}
    \caption{
        The two-stage approach for OPC recipe generation.
        The first stage employs RL to optimize OPC recipes.
        The second stage utilizes multi-modal LLM agents to efficiently summarize the results generated by RL and generate the final OPC recipes.
    }
    \label{fig:flowchart}
\end{figure*}

In this section, we present a two-stage framework, as depicted in \Cref{fig:flowchart}, with a focus on the synergistic integration of reinforcement learning (RL) and large language models (LLM) in the development of OPC recipes.
In the first stage, RL is employed to explore and identify optimal solutions for OPC recipe generation, taking into account the shapes and positions of patterns along with their surrounding contexts. Once the RL policy is trained, it can generate a set of recipes tailored to specific pattern features by adjusting EPE measurement points and fragment points.
However, the RL process is computationally intensive and impractical for full-chip applications. To overcome this limitation, the second stage harnesses the capabilities of LLM to efficiently derive recipes by synthesizing the results produced by RL. This stage involves generating a zero-shot feature pool using LLM, annotating features with a vision-language model, and constructing a decision tree. The decision tree is then utilized to produce the final OPC recipes with enhanced efficiency. This two-stage framework leverages the complementary strengths of RL and LLM to optimize OPC recipe development, striking a balance between accuracy and computational efficiency.

\input{doc/rl}

\input{doc/llm}

%% file: doc/rl.tex
\subsection{Exploration with Reinforcement Learning}
At this stage, RL can be utilized in conjunction with any OPC engines. The RL algorithm explores the parameter space and conducts a global OPC optimization, customized for various design rules and scenarios. This process aims to automate the identification of effective, fine-tuned parameter combinations, thereby reducing the time and expertise needed for manual OPC recipe adjustment.

\subsubsection{Reinforcement Learning Framework}

\input{doc/rl2.tex}

%% file: doc/rl2.tex
Proximal policy optimization (PPO)~\cite{schulman2017proximal} is an advanced RL algorithm that has shown significant potential in optimizing complex processes.
In the context of computational lithography, particularly in the development of OPC recipes,
PPO offers a powerful framework for improving the precision and efficiency of photomask patterning.
The objective of OPC is to adjust the mask design so that the printed patterns on the wafer closely match the intended design after exposure and development.
In contrast to traditional heuristic-based OPC recipe development methods~\cite{BBS-recipe-Wu-2016,asthana2016opc},
PPO leverages deep learning models to iteratively enhance OPC recipe development through agent-environment interactions.
The environment in this context includes the polygon space defined by polygon coordinates and their corresponding rasterized images,
which serve as feature embeddings in the observation space, as illustrated in \Cref{fig:flowchart}.
The agent, guided by PPO, learns to adjust fragment points and EPE measure points to minimize the OPC loss function.

The state of the environment at time $t$, denoted as $s_t$, includes both the polygon coordinates and the rasterized image features.
The agent takes an action $a_t$, which adjusts the positions of these points.
The environment then transitions to a new state $s_{t+1}$ and the agent receives a reward $r_t$ based on the OPC loss.
The goal of the PPO algorithm is to maximize the expected cumulative reward, defined as the return $R_t$:
\begin{equation}
R_t = \sum_{k=t}^{T} \gamma^{k-t} r_k,
\end{equation}
where $\gamma$ is the discount factor and $T$ is the time horizon.
PPO optimizes a policy $\pi_\theta(a_t | s_t)$, parameterized by $\theta$, by interacting with the environment and updating $\theta$ to maximize the expected return. The policy update is constrained by a proximity term to ensure stability:
\begin{equation}
  \begin{aligned}
  \mathcal{L}^{\text{CLIP}}(\theta) =
      &\mathbb{E}_{t} \Bigg[ \min \Bigg( \frac{\pi_\theta(a_t | s_t)}{\pi_{\theta_{\text{old}}}(a_t | s_t)} \hat{A}_t, \\
      &\text{clip} \left( \frac{\pi_\theta(a_t | s_t)}{\pi_{\theta_{\text{old}}}(a_t | s_t)}, 1 - \epsilon, 1 + \epsilon \right) \hat{A}_t
      \Bigg)
  \Bigg],
  \end{aligned}
\end{equation}
where $\hat{A}_t$ is the advantage estimate and $\epsilon$ is a clipping parameter.
The advantage estimate $\hat{A}_t$ is calculated as:
\begin{equation}
\hat{A}_t = \delta_t + (\gamma \lambda) \delta_{t+1} + \dots + (\gamma \lambda)^{T-t+1} \delta_T,
\end{equation}
with the temporal difference error $\delta_t$ given by:
\begin{equation}
\delta_t = r_t + \gamma V(s_{t+1}) - V(s_t).
\end{equation}

In the OPC context, the state $s_t$ includes the current positions of the measurement points, the fragment points, and the rasterized image features.
The action $a_t$ consists of permissible adjustments to these points within a specified range of $\pm 40 nm$.
The reward function $r_t$, critical to the RL training process, is derived from the OPC loss $\mathcal{L}_{\text{OPC}}$,
which quantifies the alignment between the corrected mask pattern and the target design post-lithography.
To align with the RL paradigm where higher rewards are preferred, the reward is defined as the negative of the OPC loss:
\begin{equation}
r_t = -\mathcal{L}_{\text{OPC}}.
\end{equation}
Mathematically, the OPC loss can be expressed as:
\begin{equation}
\mathcal{L}_{\text{OPC}} = \alpha \cdot \mathcal{L}_2({v}, {z}) + \beta \cdot \text{EPE}({v}, {z}) + \gamma \cdot \text{PVB}({v}, {z}),
\end{equation}
where ${v}$ represents the rasterized image representation from vertices of the polygon, ${z}$ is the target pattern, $\mathcal{L}_2$ is the Euclidean distance metric.
EPE and PVB are edge placement error and process variation band.
The coefficients $\alpha$, $\beta$, and $\gamma$ are weights that balance the contributions of each term to the overall loss.

Additionally, the value function $V(s_t)$ is approximated using a neural network parameterized by $\phi$, and is trained to minimize the following loss:
\begin{equation}
\mathcal{L}_V(\phi) = \mathbb{E}_{t} \left[ \left( V_\phi(s_t) - R_t \right)^2 \right].
\end{equation}
The overall training objective combines the clipped surrogate objective for policy optimization and the value function loss, along with an entropy bonus $S[\pi_\theta](s_t)$ to encourage exploration:
\begin{equation}
\mathcal{L}(\theta, \phi) = \mathbb{E}_{t} \left[ \mathcal{L}^{\text{CLIP}}(\theta) - c_1 \mathcal{L}_V(\phi) + c_2 S[\pi_\theta](s_t) \right],
\end{equation}
where $c_1$ and $c_2$ are coefficients that balance the importance of the value loss and the entropy bonus, respectively.

%% file: doc/llm.tex
\subsection{Recipe Summarization with LLMs}
In the second stage, LLMs are used to summarize the RL-generated OPC recipes.
By leveraging the summarization capabilities of LLMs, we derive more effective and generalizable OPC recipe rules. The use of LLMs in this stage ensures that the insights gained from the RL-generated recipes are translated into practical and applicable rules for OPC engineers.
LLMs possess advanced natural language processing capabilities~\cite{zhu2023large}, allowing them to understand and generate human-like text. This makes them well-suited for the task of summarizing~\cite{brown2020language} and reasoning~\cite{kojima2022large} over complex OPC recipes. The summarization process involves condensing the large volume of generated recipes into a coherent set of rules that can be easily understood and applied by OPC engines and engineers. The reasoning process, on the other hand, involves analyzing the summarized rules to identify patterns and relationships that can lead to the discovery of new, more effective OPC recipes.

Although the initial phase of employing RL produced better results than the baseline, it encountered significant challenges when applied to full-chip scenarios due to the need to process millions of clips, leading to an impractically long runtime for RL. Additionally, in OPC recipe development, the final step requires the extraction of recipe rules based on pattern shapes, which are then used by commercial OPC software for pattern matching and retargeting operations.

The second phase of the framework leverages the capabilities of multimodal large models to bridge the gap between superior RL exploration outcomes and the generation of OPC recipes.
To address the hallucination issues often associated with LLMs, our approach is divided into four steps:
\circled{1}
During the data processing stage, we convert the RL results into two distinct formats: JSON and image clips;
\circled{2}
We utilize LLMs to perform feature generation and zero-shot data labeling.
\circled{3}
We construct a decision tree based on the labeled features.
\circled{4}
Finally, the decision tree serves as a retrieval source for the LLM, facilitating the generation of the OPC recipe.
This structured methodology ensures the effective translation of RL exploration results into practical OPC recipes,
enhancing the overall efficiency and accuracy of the recipe generation process.

\subsubsection{Data Representation Transformation}
Since the RL action space primarily focuses on two aspects—EPE measurement movement and edge fragment point movement—the data structure can be uniformly represented in two major parts. The first part indicates whether the point's movement direction aligns with the positive direction, and the second part specifies the exact movement distance. This structured representation ensures that the RL-optimized layout information is effectively retained and utilized in OPC recipe development.
More specifically, for each point $e_i$, we will record the RL-adjusted movement vector $\boldsymbol{\delta}$ with normal direction $\boldsymbol{i}$ and the distance $\delta$.
To facilitate the LLM's understanding of the RL results, we convert the data into a JSON format.

\subsubsection{Zero-shot Feature Pool Generation and Feature Labeling}
To explicitly express the ``optimization algorithm'' embedded in the RL results, a straightforward approach is to input the RL outputs directly into the LLM as coordinates of segments.
However, this method presents two major issues: first, the LLM cannot comprehend the spatial relationships between edges or polygons based solely on coordinates; second, due to the limitations of the LLM's context window, it cannot process exceedingly long coordinate representations, leading to judgments based only on the first few points, often resulting in incorrect assessments.
To maximally preserve the layout information optimized by RL, we transcribed the information, recording location-related details and geometry features for each point.

In traditional OPC recipe development, engineers manually set EPE measurement points and fragment points based on pattern shape, position, surrounding pattern characteristics, and layer number. This is followed by forward lithography and OPC simulation to obtain evaluation results, which are iteratively adjusted. This process is time-consuming, labor-intensive, and monotonous.
Recent studies, such as \cite{GPTlaber-Fabrizio-2023}, have demonstrated that LLMs outperform human annotators in tasks related to labeling and annotation, offering higher efficiency and lower costs. Automating this process with LLMs is a straightforward idea that can significantly enhance engineers' efficiency.
Our method of utilizing LLMs for classification labeling in OPC recipe development involves two steps: first, feature pool mining and generation. Second, feature labeling.

\subsubsection{Feature Mining and Generation}
Raw images of EPE points and fragment points are input into a multi-modality large language model (MLLM), which analyzes the images and extracts features. For different points, the features generated by the LLM are pooled and deduplicated. The data format includes feature names and descriptions, resulting in a comprehensive feature pool.
Part of the feature pool is shown below and the full feature pool is available in the supplementary material.
Those features will be fed into next step for the labeling agents to label the features for each EPE measurement point and fragment point.
\begin{tcolorbox}[title={Feature pool examples}]
  \small
  \{\texttt{on\allowbreak\_jog\allowbreak\_long\allowbreak\_edge}: ``it is on the jog, but on the long edge of the jog'',\\
  \texttt{on\allowbreak\_jog\allowbreak\_short\allowbreak\_edge}: ``it is on the jog, but on the short edge of the jog'',\\
  \texttt{on\allowbreak\_horizontal\allowbreak\_edge}: ``the point is located on a horizontal edge'',\\
  \texttt{on\allowbreak\_vertical\allowbreak\_edge}: ``the point is located on a vertical edge'',
  \dots\}
\end{tcolorbox}

\subsubsection{Feature Labeling}
Using the feature pool generated by MLLMs in the first step, for each EPE measurement point and fragment point, we employ the MLLMs to label the input images with corresponding prompts and evaluate each feature in the pool.
By labeling each point, we obtain a series of feature information for each point, which is then used to construct our decision tree. This automated approach not only preserves the intricate details of the RL-optimized layouts but also streamlines the OPC recipe development process, making it more efficient and effective.
Another critical aspect is the labeling of ground truth for decision trees.
To ensure the recipe does not become overly complicated, the RL-movement vector $\boldsymbol{\delta}$ is divided into different intervals.
These intervals serve as classification boundaries.
For example, in a given direction, the vector is categorized into $C$ intervals, where the furthest positive interval is labeled as $+C$ and the furthest negative interval as $-C$.
Consequently, the ground truth labels range from $-C$ to $+C$, encompassing a total of $2C + 1$ categories.
A label of $0$ indicates no movement.
\begin{tcolorbox}[title = {LLM Labeling Example}]
  \small
  \{\texttt{epe\allowbreak\_id}: 10, \\
  \texttt{features} : \{
      \texttt{on\allowbreak\_jog\allowbreak\_long\allowbreak\_edge}: false,
      \texttt{on\allowbreak\_jog\allowbreak\_short\allowbreak\_edge}: false,
      \texttt{on\allowbreak\_horizontal\allowbreak\_edge}: true,
      \texttt{on\allowbreak\_vertical\allowbreak\_edge}: false,
      \dots \},
  \texttt{result}: $+C$\}\} \dots \\
  \{\texttt{epe\allowbreak\_id}: 15, \\
  \texttt{features} : \{
      \texttt{on\allowbreak\_jog\allowbreak\_long\allowbreak\_edge}:, true
      \texttt{on\allowbreak\_jog\allowbreak\_short\allowbreak\_edge}: false,
      \texttt{on\allowbreak\_horizontal\allowbreak\_edge}: false,
      \texttt{on\allowbreak\_vertical\allowbreak\_edge}: true,
      \dots \},
  \texttt{result}: $-C$\}\}
\end{tcolorbox}

\subsubsection{Self-improvement System}
\begin{figure}[tb!]
  \centering
  \includegraphics[width=0.96\linewidth]{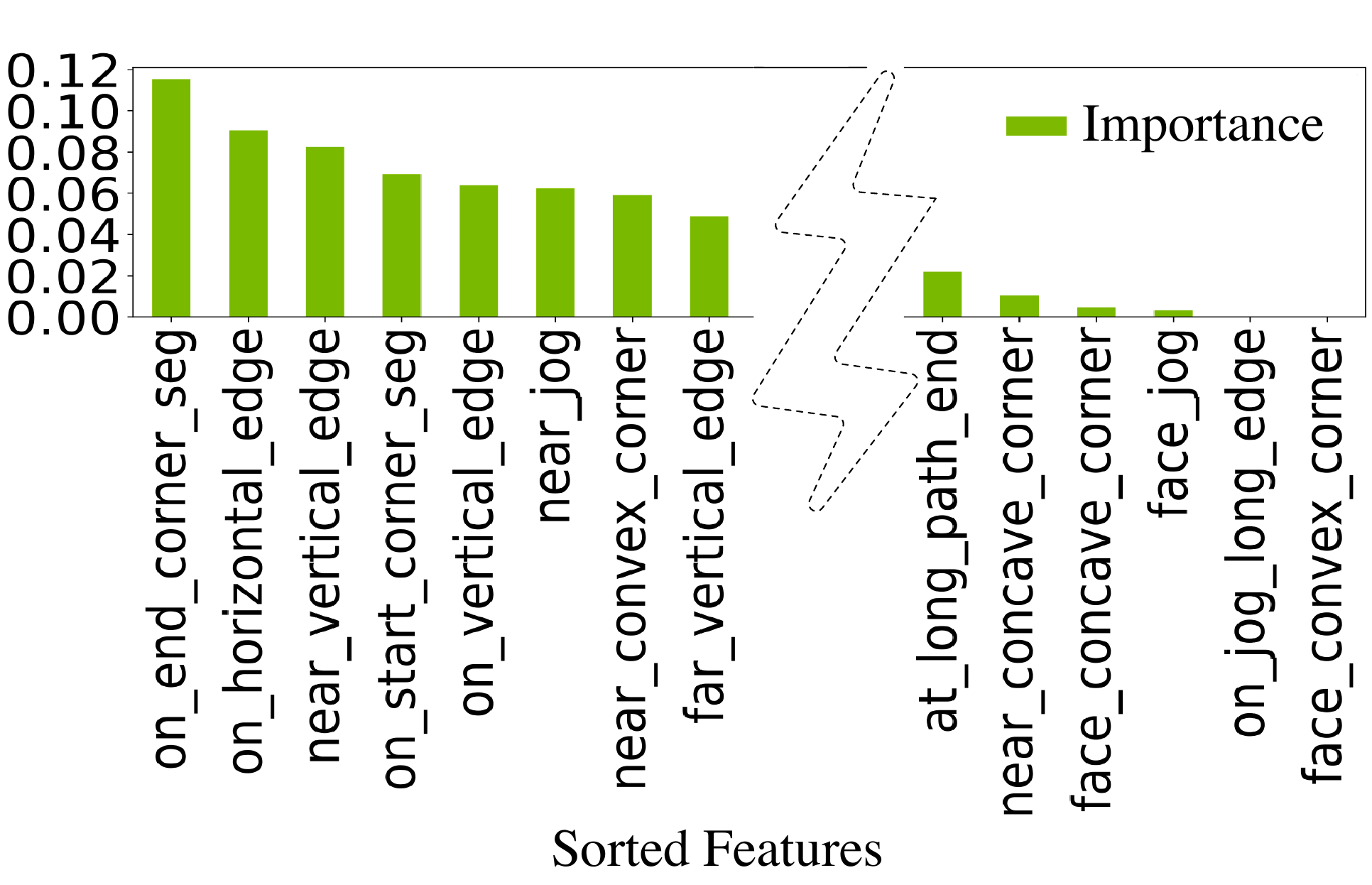}
  \caption{
      Feature importance ranking.
      The features are ranked based on their importance in the decision tree.
      The less important features are removed and fed back into the LLM for further improvement.
  }
  \label{fig:llm-feature-importance}
\end{figure}

Upon generating the decision tree, we can rank the features by importance and perform importance-based feature selection.
As illustrated in \Cref{fig:llm-feature-importance}, the features \texttt{on jog long edge} and \texttt{face convex corner} rank the lowest, with a feature importance of 0.
By removing the unimportant features and recycling them back into the LLM's input, we update the feature pool and continue constructing new decision trees. This iterative process is repeated several times, enhancing feature extraction and development, ultimately making the system self-improving.

This approach mirrors the workflow of human OPC engineers, who continuously experiment, validate results, and refine recipes. By automating this process into a symbolic improvement and generation system based on decision trees, we streamline and enhance the traditional OPC recipe development process. This automation not only accelerates the creation of optimized OPC recipes but also improves their accuracy and reliability, demonstrating the significant potential of integrating advanced computational techniques into the domain of computational lithography.

\subsubsection{Recipe Generation Based on Decision Tree}
\begin{figure}[!t]
  \centering
  \subfloat[Decision tree example. The leaf nodes of the decision tree are labeled with ground truth based on RL results. The non-leaf nodes are features generated and labeled by the LLM.\label{fig:llm-recipe-tree}]{
    \includegraphics[width=0.99\linewidth]{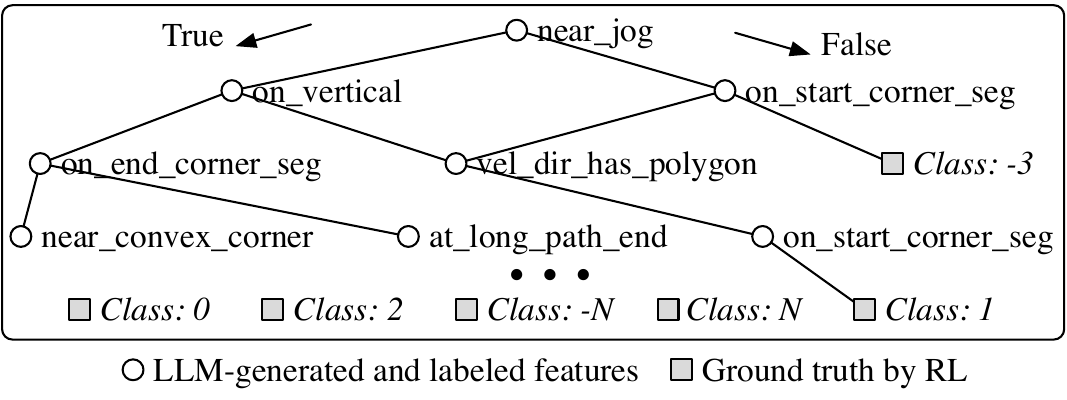}
  }\\
  \subfloat[LLM-generated recipe example in \texttt{jsonl} format. Conditions are tree feature labels, and the type determines the tasks. The class represents the RL ground truth.
  \label{fig:llm-recipe}]{
    \begin{tcolorbox}[colback=gray!15!white,boxsep=1mm,boxrule=0mm,frame hidden,width=\linewidth,left=3mm,right=3mm]
      \dots\{\texttt{condition}: [\texttt{near\allowbreak\_jog, on\allowbreak\_vertical, not vel\allowbreak\_dir\allowbreak\_has\allowbreak\_polygon, not on\allowbreak\_start\allowbreak\_corner\allowbreak\_seg}\ ], \texttt{type}: \texttt{EPE}, \texttt{class}: 1\}
      \{\texttt{condition}: [\texttt{not near\allowbreak\_jog, not on\allowbreak\_start\allowbreak\_corner\allowbreak\_seg}\ ], \texttt{type}: \texttt{EPE}, \texttt{class}: 3\}
      \{\texttt{condition}: [\texttt{near\allowbreak\_convex\allowbreak\_corner, next\allowbreak\_to\allowbreak\_concave\allowbreak\_corner}\ ], \texttt{type}: \texttt{FRAG}, \texttt{class}: -2\} \dots
    \end{tcolorbox}
  }\\
  \subfloat[Downstream OPC software recipe example also includes statements for defining feature labels and movement distances.\label{fig:llm-recipe-calibre}]{
    \begin{tcolorbox}[colback=gray!15!white,boxsep=1mm,boxrule=0mm,frame hidden,width=\linewidth,left=3mm,right=3mm]
      \texttt{NEWTAG edge A short\allowbreak\_len corner1 convex corner2 convex -out line\allowbreak\_end}\\
      \texttt{NEWTAG neighbor both line\allowbreak\_end short\allowbreak\_len corner convex -out line\allowbreak\_end\allowbreak\_adj}\\
      \texttt{fragment\allowbreak\_corner A convex concave mid\allowbreak\_length 0.03}\\
      \texttt{fragment\allowbreak\_corner convex concave long\allowbreak\_lenght 0.04 breakinhalf}\\
      \texttt{retarget\allowbreak\_layer A pattern0 curve0 pattern\allowbreak\_epe curve1 emulate}
    \end{tcolorbox}
  }
  \caption{
      Decision tree and recipe generation example.
      (a) Decision tree constructed from LLM-labeled features and RL ground truth.
      (b) Simplified LLM-generated recipe example in \texttt{jsonl} format.
      (c) Downstream OPC~\cite{TOOL-calibre-OPC} software recipe example in Tcl language.
  }
  \label{fig:main-figure}
\end{figure}

After annotating with the LLM, we can intuitively construct a decision tree for each pattern by combining the original polygon information with the extracted features.
Based on the results from the first RL phase, we label each leaf node of the decision tree with the corresponding recipe type.
Once the decision tree is trained, each branch serves as a reference for the LLM to write the corresponding rules, ultimately generating a complete OPC recipe, as shown in \Cref{fig:llm-recipe-tree} and \Cref{fig:llm-recipe}.
Additionally, using the decision tree as an intermediate representation has the benefit that the LLM can, after abstracting the decision tree, generate different recipe expressions according to the syntax of various downstream OPC software, as shown in \Cref{fig:llm-recipe-calibre}.

%% file: doc/exp.tex
\section{Experimental Results}
\label{sec:exp}
\subsection{Dataset}
To evaluate the effectiveness of our framework, we utilized datasets from two distinct processes.
The first dataset is derived from the 2013 ICCAD contest~\cite{OPC-ICCAD2013-Banerjee},
which includes a test set of ten $2\mu m \times 2\mu m$ metal layer patterns fabricated using a 32nm process.
This dataset is widely employed in various OPC engines and semiconductor lithography research.
We utilized the training set provided by~\cite{OPC-TCAD2020-Yang} for our experiments.
The second dataset is sourced from the NVIDIA Deep Learning Accelerator (NVDLA)~\cite{NVDLA},
an open architecture designed to standardize deep learning inference accelerators.
From the full-chip layout of the NVDLA, fabricated using NanGate 45$nm$ standard cells~\cite{FreePDK45}, we extracted nearly one million clips.
We then randomly selected 800 clips for the training set and 200 clips for the test set.
This diverse dataset collection enabled a comprehensive evaluation of our OPC recipe development and its application within computational lithography.

\subsection{Model and OPC Engine}
In this study, we utilize GPT-4o~\cite{gpt-4o}, an optimized version of GPT-4 with multi-modal capabilities that process both text and images,
enhancing performance and versatility.
The feature labeling component relies on these multi-modal capabilities, while other parts of the framework can use a purely language-based model.
The OPC model is built on top of the open-source OPC engine \cite{OPC:OpenILT}, which is widely used in the OPC community.
The methodology can be applied to any OPC engine.
Implementation details and prompt scripts are provided in the supplementary material.
The hyperparameters of the OPC loss are set to $\alpha=1$, $\beta=100$, and $\gamma=1$.

\subsection{Decision Tree Efficiency}
\begin{figure}[tb!]
  \centering
  \includegraphics[width=0.99\linewidth]{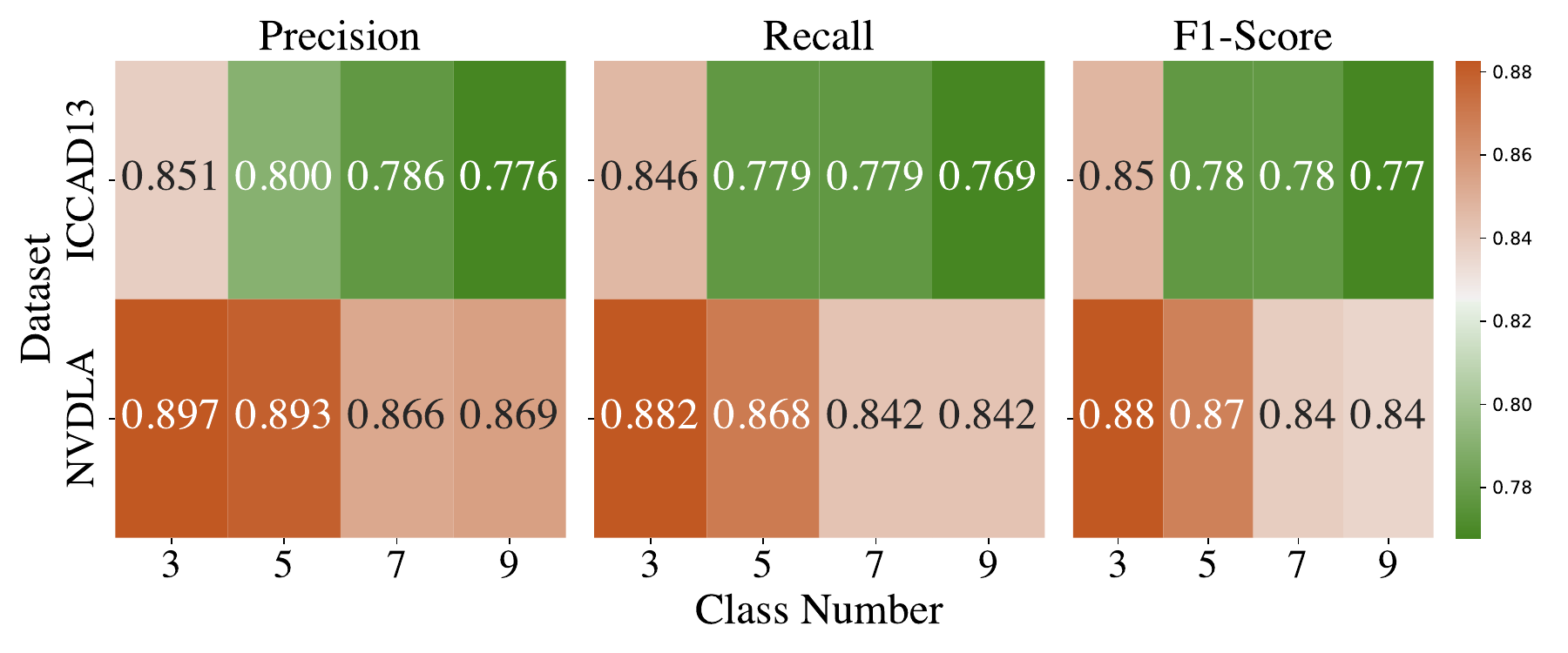}
  \caption{Decision Tree Efficiency on both ICCAD13 and NVDLA datasets.}
  \label{fig:decision_tree_efficiency}
\end{figure}

As a critical basis for generating recipes using LLMs, decision tree models indirectly influence the final OPC performance.
In our study, we demonstrate the precision, recall, and F1-score of the final decision tree generated for ICCAD13 and NVDLA datasets in \Cref{fig:decision_tree_efficiency}.
The horizontal axis in \Cref{fig:decision_tree_efficiency} represents the number of segments into which we divided the displacement distance, indicating the number of classes.
The average precision ranges from 0.77 to 0.90, with some examples achieving a precision of 100\%.
This indicates that the decision tree constructed using features automatically mined by the LLM agent can achieve 100\% accuracy on certain patterns.
However, as the complexity and number of patterns increase, the overall average precision falls below 0.9. 
Despite this, for overall OPC effectiveness, a higher number of classes results in more precise outcomes. Therefore, in the recipe optimized by the LLM,
we utilized 9 classes to compare the OPC results.

\subsection{Efficiency of the Overall Framework}

\input{doc/res/all.tex}

In \Cref{tab:all}, we present the results of three different scenarios: the pure OPC engine~\cite{OPC:OpenILT}, the OPC engine after first-stage RL optimization, and the OPC engine utilizing an LLM-generated recipe.
The metrics PVB, EPE N, and EPE D are introduced before, with smaller values indicating better performance.
The runtime is measured in seconds for the OPC engine and in hours for the RL optimization.
In both the ICCAD13 and NVDLA benchmarks, RL optimization significantly reduces EPE values.
Specifically, in the ICCAD13 benchmark, RL reduces the EPE number by 12\% and the EPE distance by nearly 24\%.
Similarly, in the NVDLA benchmark, RL reduces the EPE number by 13\% and the EPE distance by 15\%.
However, RL's major drawback is its excessive time consumption for pattern optimization,
highlighting the necessity of the second stage of our framework.

The results of the second stage, shown in the OPC+LLM column of the table, indicate that the LLM-generated recipe performs comparably to RL optimization.
For the NVDLA benchmark, the LLM-generated recipe reduces EPE N and EPE D by 8\% and 12\%,
while for the ICCAD13 benchmark, it reduces EPE N by 11\% and EPE D by 19\%.
The advantage of the rule-based LLM-generated recipe is its immediate applicability to new layouts without additional optimization time, maintaining the same runtime as the OPC engine itself while enhancing performance.

%% file: doc/res/all.tex
\begin{table}[!tbp]
    \centering
    \caption{Performance of the framework. The RL stage results are shown in the OPC+RL column, and the results related to the final LLM-generated recipe are shown in the OPC+LLM column.}
    \label{tab:all}
  \setlength{\tabcolsep}{6pt}
  \renewcommand{\arraystretch}{1.0}
  \begin{tabular}{l|lll}\toprule
  &\multicolumn{3}{c}{ICCAD13} \\
  &OPC &OPC+LLM &OPC+RL \\\midrule
  PVBand &53328 &51271 &50060 \\
  \cellcolor[HTML]{f3f3f3}ratio &\cellcolor[HTML]{f3f3f3}1.00 &\cellcolor[HTML]{f3f3f3}0.96 &\cellcolor[HTML]{f3f3f3}0.94 \\
  EPE N &119.70 &107.00 &105.60 \\
  \cellcolor[HTML]{f3f3f3}ratio &\cellcolor[HTML]{f3f3f3}1.00 &\cellcolor[HTML]{f3f3f3}0.89 &\cellcolor[HTML]{f3f3f3}0.88 \\
  EPE D &693.10 &561.70 &525.90 \\
  \cellcolor[HTML]{f3f3f3}ratio &\cellcolor[HTML]{f3f3f3}1.00 &\cellcolor[HTML]{f3f3f3}0.81 &\cellcolor[HTML]{f3f3f3}0.76 \\
  Runtime &4s &4s &3hr \\\midrule\midrule
  &\multicolumn{3}{c}{NVDLA} \\
  &OPC &OPC+LLM &OPC+RL \\\midrule
  PVBand &170899 &161056 &162096 \\
  \cellcolor[HTML]{f3f3f3}ratio &\cellcolor[HTML]{f3f3f3}1.00 &\cellcolor[HTML]{f3f3f3}0.94 &\cellcolor[HTML]{f3f3f3}0.95 \\
  EPE N &159.45 &146.95 &139.40 \\
  \cellcolor[HTML]{f3f3f3}ratio &\cellcolor[HTML]{f3f3f3}1.00 &\cellcolor[HTML]{f3f3f3}0.92 &\cellcolor[HTML]{f3f3f3}0.87 \\
  EPE D &869.25 &766.10 &741.05 \\
  \cellcolor[HTML]{f3f3f3}ratio &\cellcolor[HTML]{f3f3f3}1.00 &\cellcolor[HTML]{f3f3f3}0.88 &\cellcolor[HTML]{f3f3f3}0.85 \\
  Runtime &6s &6s &3hr \\
  \bottomrule
  \end{tabular}
  \end{table}

%% file: doc/conclu.tex
\section{Conclusion}
\label{sec:conclusion}
In this paper, we propose a two-stage framework to automate the task of OPC recipe development.
In the first stage, RL mimics the process by which an OPC engineer designs a recipe.
This involves exploring the optimal solutions for EPE measurement and fragmentation based on the characteristics of the patterns.
In the second stage, LLMs automate the process of summarizing the recipe crafted by the OPC engineer.
Utilizing the optimal solutions generated by RL, the LLM initially generates a relevant feature pool from the layout and subsequently annotates each point with its corresponding features.
This annotated data is then used in conjunction with the RL results to construct a decision tree.
Ultimately, the LLM retrieves this decision tree to generate a set of recipes.
The experimental results show that this framework reduces the key error metric by more than 10\% without increasing runtime.

%% file: doc/appendix.tex
\clearpage
\twocolumn
\appendix
\section{Appendix}
\subsection{More background on optical proximity correction (OPC)}

OPC is a pivotal technology in the domain of computational lithography, designed to counteract the optical proximity effects (OPE) that emerge from diffraction and interference during the photolithographic process. As semiconductor manufacturing has advanced, particularly with the reduction of technology nodes to 90 nm and below, the limitations of conventional resolution enhancement techniques (RET) have become increasingly apparent. This has driven the evolution of OPC from rudimentary approaches to sophisticated methodologies capable of addressing the complex challenges of modern lithography.

\begin{figure}[h]
  \centering
  \includegraphics[width=0.96\linewidth]{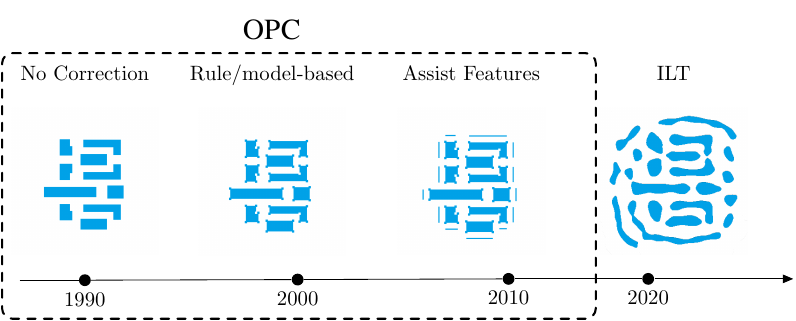}
  \caption{The evolution of OPC technology.}
  \label{fig:timeline}
\end{figure}

As illustrated in \Cref{fig:timeline}, the initial phase of OPC, developed in the early 1990s, was predominantly rule-based (RBOPC). This technique employed predetermined correction rules derived from empirical observations and simulations to correct local OPEs. While RBOPC was computationally efficient and straightforward, it lacked the capacity to optimize mask patterns on a global scale, which became increasingly necessary as device geometries shrank and layouts grew more complex.

In response to the limitations of RBOPC, the early 2000s saw the advent of Model-Based OPC (MBOPC). MBOPC introduced a more advanced approach by leveraging detailed physical models of the lithographic process and applying numerical optimization techniques to refine mask patterns. This paradigm shift allowed for a more precise and globally optimized correction of OPEs. Within MBOPC, two primary approaches emerged: Edge-Based OPC (EBOPC) and Pixel-Based OPC (PBOPC), also known as Inverse Lithography Technology (ILT).

EBOPC focuses on optimizing the edges of mask patterns by iteratively adjusting the positions of edge segments to minimize imaging errors. Although EBOPC offers improved accuracy over RBOPC, it faces significant challenges in terms of computational efficiency, particularly when dealing with the complex interactions and nonlinearities present in advanced lithography processes.

PBOPC or ILT, in contrast, represents a more granular approach by treating the mask as a pixelated grid, where each pixel's transmission is optimized. This allows for highly complex mask patterns, including the creation of free-form shapes and sub-resolution assist features (SRAF) that can significantly improve imaging performance. However, the increased complexity of ILT-generated masks presents challenges in manufacturing, particularly in the conversion of these intricate patterns into manufacturable layouts. The need for rectangular decomposition and rigorous mask rule checks (MRC) often introduces new complexities and potential hotspots, which can negate some of the performance gains achieved through ILT.

In the 2010s and beyond, ILT has continued to evolve, with the development of advanced algorithms such as Multi-Level Inverse Lithography Technology (MultiILT). These newer methods aim to strike a balance between performance and manufacturability by employing multi-level resolution strategies. However, despite these advancements, ILT and other pixel-based methods are still not widely adopted in the industry due to the high complexity and cost associated with their implementation.

OPC recipe development is a vital and intricate process in semiconductor manufacturing, essential for configuring OPC software to correct for optical distortions and proximity effects during lithography.
As semiconductor devices scale down to sub-10 nm nodes, the complexity of integrated circuit layouts increases, making precise OPC recipe development crucial for ensuring accurate pattern transfer. These recipes involve setting parameters and special handling cases within the OPC tools to optimize the correction for different geometries and process variations, balancing between optimal optical correction and manufacturability. The process is challenging due to the need for extensive iteration, rigorous testing, and the consideration of process variability, making OPC recipe development a time-consuming and resource-intensive task, yet indispensable for achieving high-yield and high-performance semiconductor devices.

\subsection{Prompts and Feature Pools}
The prompt used for \Cref{sec:method} in zero-shot feature pool generation and feature mining \& labeling is listed below, including the task description, feature characteristics, final result format, and more.
\begin{tcolorbox}[title={Feature Mining Prompt},breakable]
\small
In the image I give to you, this is a layout image depicting the lithography and OPC process in the semiconductor industry.
Within the image, there is a red dot indicating the EPE measurement point.
Could you please identify and describe the features surrounding that red dot as much as possible (location features, relation features, geometry features, proximity features).
I will provide you with an example feature list  format and their definitions, please give me more binary features (true or false, feature number at least 15) and their definitions.
Remove duplicate features if they can be merged by the value.
and request your response in JSON format.

\{
  features: \{
      ``types'': ``CV'',\\
      ``near\allowbreak\_vel\allowbreak\_dir\allowbreak\_has\allowbreak\_polygon'': false,\\
      ``far\allowbreak\_vel\allowbreak\_dir\allowbreak\_has\allowbreak\_polygon'': false,\\
      ``near\allowbreak\_jog'': false,\\
      ``on\allowbreak\_start\allowbreak\_corner\allowbreak\_seg'': true,\\
      ``on\allowbreak\_end\allowbreak\_corner\allowbreak\_seg'': false\\
  \},\\
  explain: {
      ``types: ''CV for the corner vertical, CH for the corner horizontal'',\\
      ``near\allowbreak\_vel\allowbreak\_dir\allowbreak\_has\allowbreak\_polygon'': ''in the near distance , there are polygon faced on that edge where it located on'',\\
      ``far\allowbreak\_vel\allowbreak\_dir\allowbreak\_has\allowbreak\_polygon'': ''in the far distance, there are polygon faced on that edge there it located on'',\\
      ``near\allowbreak\_jog'': ''in the near distance, there is a jog on that edge where it located on'',\\
      ``on\allowbreak\_start\allowbreak\_corner\allowbreak\_seg'': ''it is on the corner start segment by clock wise'',\\
      ``on\allowbreak\_end\allowbreak\_corner\allowbreak\_seg'': ''it is on the corner end segment by clock wise''\\
  }
\}
\end{tcolorbox}

The final generated feature pools are listed below:
\begin{tcolorbox}[title={The LLM generated feature pools.},breakable]
\small
\{
    ``types'': ``CV for the corner vertical, CH for the corner horizontal, H for the horizontal but not on corner, V for the vertical but not on corner.'', \\
    ``near\_jog'': ``in the near distance, there is a jog on that edge where it is located'', \\
    ``face\_jog'': ``there is a jog facing the point'', \\
    ``on\_jog\_long\_edge'': ``it is on the jog, but on the long edge of the jog'', \\
    ``on\_jog\_short\_edge'': ``it is on the jog, but on the short edge of the jog'', \\
    ``on\_start\_corner\_seg'': ``it is on the corner start segment by clockwise'', \\
    ``on\_end\_corner\_seg'': ``it is on the corner end segment by clockwise'', \\
    ``near\_hor\_dir\_has\_polygon'': ``in the near horizontal direction, there are polygons facing on that edge where it is located, but not connected on edge'', \\
    ``far\_hor\_dir\_has\_polygon'': ``in the far horizontal direction, there are polygons facing on that edge where it is located, but not connected on edge'', \\
    ``near\_ver\_dir\_has\_polygon'': ``in the near vertical direction, there are polygons facing on that edge where it is located, but not connected on edge'', \\
    ``far\_ver\_dir\_has\_polygon'': ``in the far vertical direction, there are no polygons facing on that edge where it is located, but not connected on edge'', \\
    ``on\_horizontal\_edge'': ``the point is located on a horizontal edge'', \\
    ``on\_vertical\_edge'': ``the point is located on a vertical edge'', \\
    ``near\_convex\_corner'': ``there is a convex corner near the point, connected on edge'', \\
    ``near\_concave\_corner'': ``there is a concave corner near the point, connected on edge'', \\
    ``face\_convex\_corner'': ``there is a convex corner facing the point, not connected on edge'', \\
    ``face\_concave\_corner'': ``there is a concave corner facing the point, not connected on edge'', \\
    ``near\_horizontal\_edge'': ``there is a horizontal edge near the point'', \\
    ``near\_vertical\_edge'': ``there is a vertical edge near the point'', \\
    ``far\_horizontal\_edge'': ``there is a horizontal edge far from the point'', \\
    ``far\_vertical\_edge'': ``there is a vertical edge far from the point'', \\
    ``at\_long\_path\_end'': ``the point is located at the end of a long path'', \\
    ``at\_short\_path\_end'': ``the point is located at the end of a short path'', \\
    ``at\_long\_path\_side'': ``the point is located at the side of a long path'', \\
    ``at\_short\_path\_side'': ``the point is located at the side of a short path'' \\
\}
\end{tcolorbox}

\subsection{Dataset samples and visualization of results}

\begin{figure}[h]
  \centering
  \includegraphics[width=0.9\linewidth]{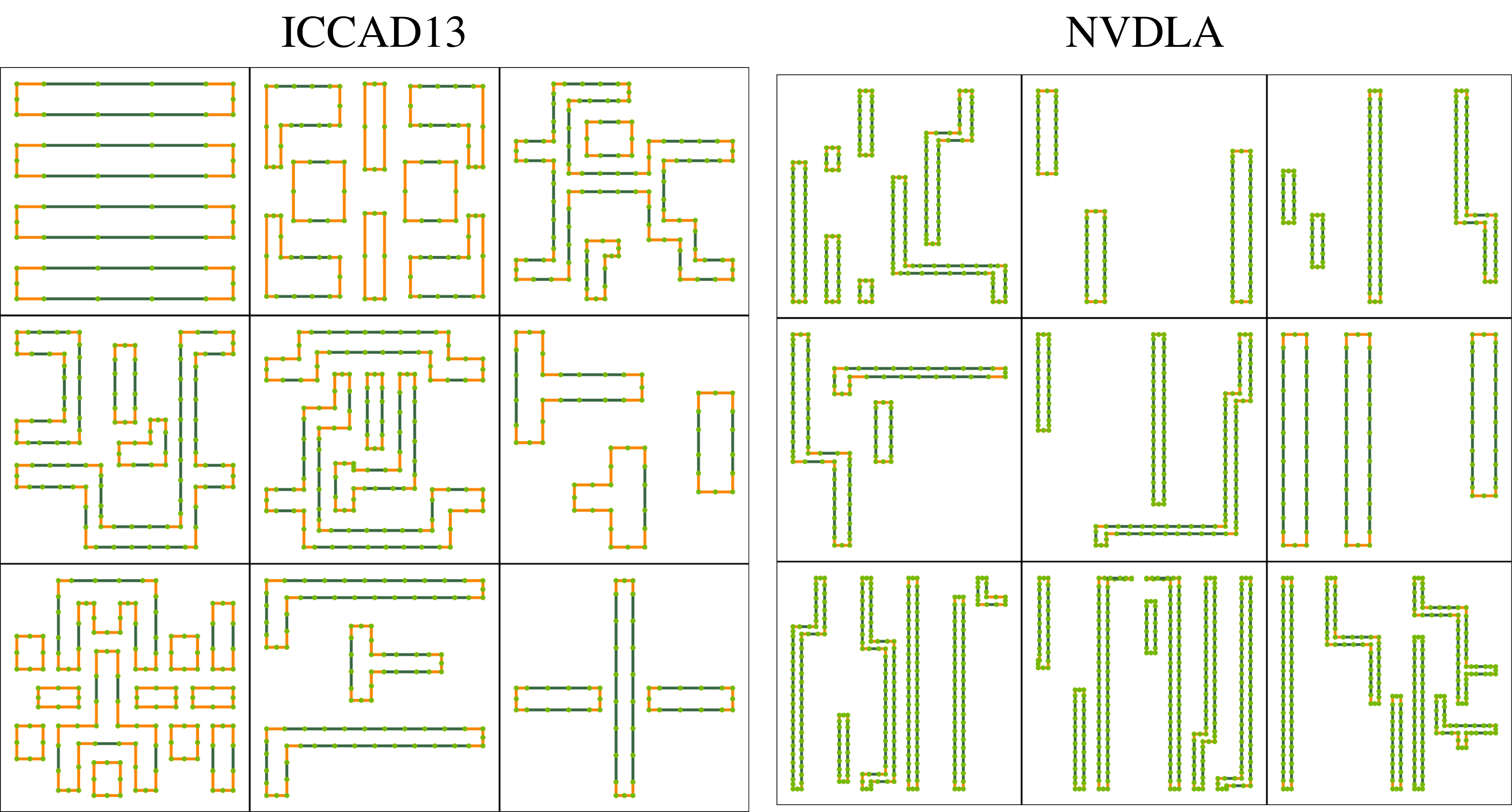}
  \caption{Samples of the ICCAD13 and NVDLA dataset.}
  \label{fig:dataset-samples}
\end{figure}

\begin{figure}[h]
  \centering
  \includegraphics[width=0.8\linewidth]{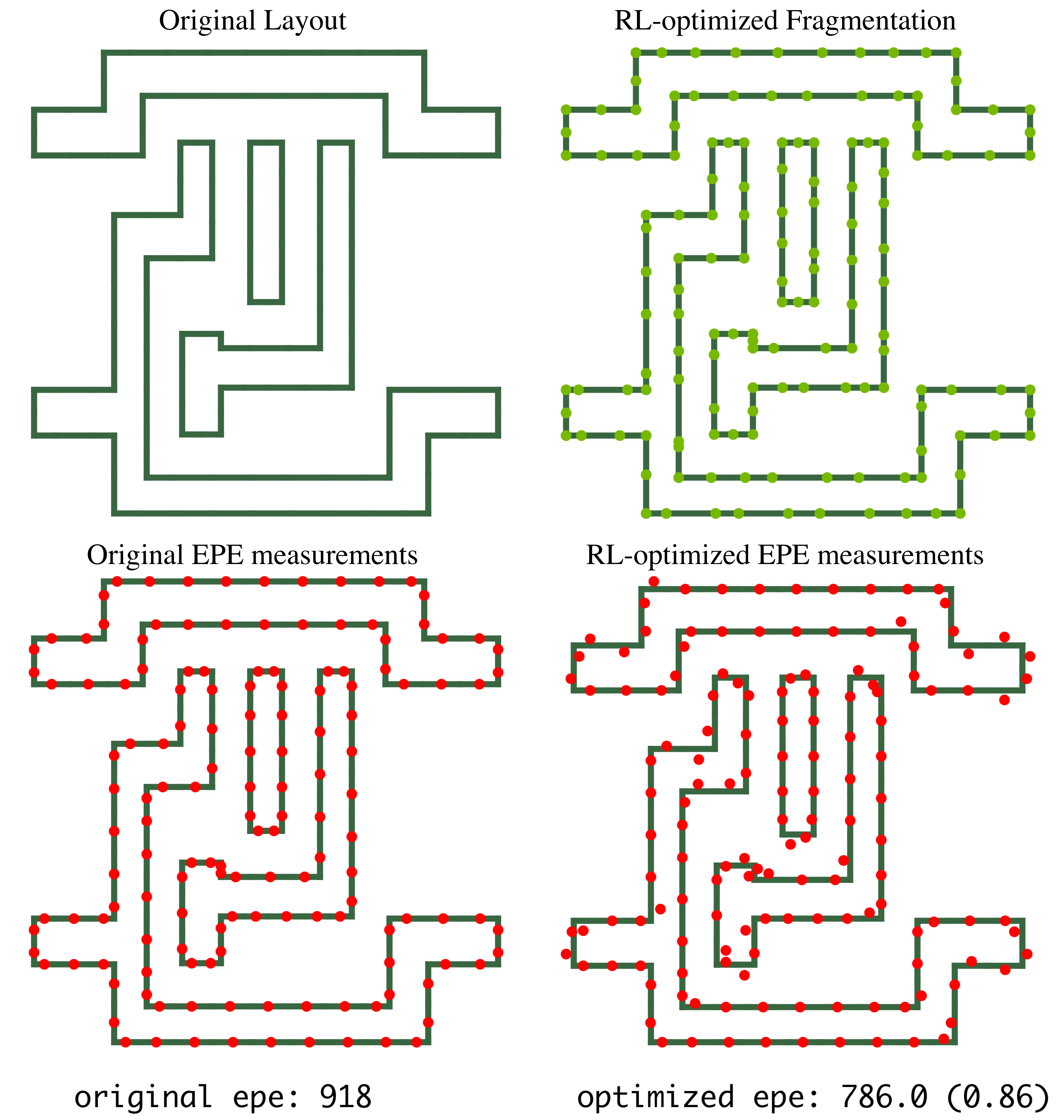}
  \caption{Visualization of the RL-optimized recipe results.}
  \label{fig:rl-results}
\end{figure}

In the \Cref{fig:dataset-samples}, we illustrate samples from both ICCAD13 and NVDLA datasets to validate the algorithm's effectiveness and robustness.
Moreover, we present the original layout of a specific example in \Cref{fig:rl-results}, the fragmentation points following RL optimization, and the EPE measurement points, as well as the corresponding reduction in EPE distance.

%% file: main.bbl
\begin{thebibliography}{26}
\providecommand{\natexlab}[1]{#1}

\bibitem[{Asthana, Wilkinson, and Power(2016)}]{asthana2016opc}
Asthana, A.; Wilkinson, B.; and Power, D. 2016.
\newblock OPC recipe optimization using genetic algorithm.
\newblock In \emph{Optical Microlithography XXIX}, volume 9780, 43--54. SPIE.

\bibitem[{Banerjee, Li, and Nassif(2013)}]{OPC-ICCAD2013-Banerjee}
Banerjee, S.; Li, Z.; and Nassif, S.~R. 2013.
\newblock {ICCAD-2013 CAD} contest in mask optimization and benchmark suite.
\newblock In \emph{IEEE/ACM International Conference on Computer-Aided Design
  (ICCAD)}, 271--274.

\bibitem[{Brown et~al.(2020)Brown, Mann, Ryder, Subbiah, Kaplan, Dhariwal,
  Neelakantan, Shyam, Sastry, Askell et~al.}]{brown2020language}
Brown, T.; Mann, B.; Ryder, N.; Subbiah, M.; Kaplan, J.~D.; Dhariwal, P.;
  Neelakantan, A.; Shyam, P.; Sastry, G.; Askell, A.; et~al. 2020.
\newblock Language models are few-shot learners.
\newblock \emph{Annual Conference on Neural Information Processing Systems
  (NeurIPS)}, 33: 1877--1901.

\bibitem[{Chen et~al.(2020)Chen, Chen, Ma, Yang, and Yu}]{OPC-ICCAD2020-DAMO}
Chen, G.; Chen, W.; Ma, Y.; Yang, H.; and Yu, B. 2020.
\newblock {DAMO}: Deep Agile Mask Optimization for Full Chip Scale.
\newblock In \emph{IEEE/ACM International Conference on Computer-Aided Design
  (ICCAD)}.

\bibitem[{Gilardi, Alizadeh, and Kubli(2023)}]{GPTlaber-Fabrizio-2023}
Gilardi, F.; Alizadeh, M.; and Kubli, M. 2023.
\newblock ChatGPT outperforms crowd workers for text-annotation tasks.
\newblock \emph{Proceedings of the National Academy of Sciences}, 120(30).

\bibitem[{Granik and Cobb(2002)}]{MEEF-Photomask2002-Granik}
Granik, Y.; and Cobb, N.~B. 2002.
\newblock {MEEF} as a Matrix.
\newblock In \emph{Photomask}, 980--991.

\bibitem[{Kojima et~al.(2022)Kojima, Gu, Reid, Matsuo, and
  Iwasawa}]{kojima2022large}
Kojima, T.; Gu, S.~S.; Reid, M.; Matsuo, Y.; and Iwasawa, Y. 2022.
\newblock Large language models are zero-shot reasoners.
\newblock \emph{Advances in neural information processing systems}, 35:
  22199--22213.

\bibitem[{Lei et~al.(2014)Lei, Hong, Lippincott, and Word}]{MEEF-lei2014model}
Lei, J.; Hong, L.; Lippincott, G.; and Word, J. 2014.
\newblock Model-based OPC using the MEEF matrix II.
\newblock In \emph{Optical Microlithography XXVII}, volume 9052, 170--178.
  SPIE.

\bibitem[{Liang et~al.(2023)Liang, Ouyang, Yang, Yu, and Ma}]{liang2023rl}
Liang, X.; Ouyang, Y.; Yang, H.; Yu, B.; and Ma, Y. 2023.
\newblock {RL-OPC}: Mask Optimization with Deep Reinforcement Learning.
\newblock \emph{IEEE Transactions on Computer-Aided Design of Integrated
  Circuits and Systems}.

\bibitem[{Liang et~al.(2024)Liang, Yang, Liu, Yu, and Ma}]{liang2024camo}
Liang, X.; Yang, H.; Liu, K.; Yu, B.; and Ma, Y. 2024.
\newblock CAMO: Correlation-Aware Mask Optimization with Modulated
  Reinforcement Learning.
\newblock In \emph{ACM/IEEE Design Automation Conference (DAC)}.

\bibitem[{Liu et~al.(2023)Liu, Ene, Kirby, Cheng, Pinckney, Liang, Alben,
  Anand, Banerjee, Bayraktaroglu et~al.}]{liu2023chipnemo}
Liu, M.; Ene, T.-D.; Kirby, R.; Cheng, C.; Pinckney, N.; Liang, R.; Alben, J.;
  Anand, H.; Banerjee, S.; Bayraktaroglu, I.; et~al. 2023.
\newblock {ChipNeMo}: Domain-adapted llms for chip design.
\newblock \emph{arXiv preprint arXiv:2311.00176}.

\bibitem[{Liu and Zhang(2010)}]{qingwei-opc-cost-2010}
Liu, Q.; and Zhang, L. 2010.
\newblock {Optimize the OPC control recipe with cost function}.
\newblock In \emph{Photomask Technology 2010}, volume 7823. SPIE.

\bibitem[{{Mentor Graphics, Siemens}(2024)}]{TOOL-calibre-OPC}
{Mentor Graphics, Siemens}. 2024.
\newblock Calibre Computational Lithography.
\newblock
  \url{https://eda.sw.siemens.com/en-US/ic/calibre-manufacturing/computational-lithography/}.

\bibitem[{{NVIDIA}(2024)}]{NVDLA}
{NVIDIA}. 2024.
\newblock NVIDIA Deep Learning Accelerator.
\newblock \url{https://nvdla.org/index.html}.

\bibitem[{{OpenAI}(2024)}]{gpt-4o}
{OpenAI}. 2024.
\newblock Hello gpt-4o.
\newblock \url{https://openai.com/index/hello-gpt-4o.}

\bibitem[{Schulman et~al.(2017)Schulman, Wolski, Dhariwal, Radford, and
  Klimov}]{schulman2017proximal}
Schulman, J.; Wolski, F.; Dhariwal, P.; Radford, A.; and Klimov, O. 2017.
\newblock Proximal policy optimization algorithms.
\newblock \emph{arXiv preprint arXiv:1707.06347}.

\bibitem[{Stine et~al.(2007)Stine, Castellanos, Wood, Henson, Love, Davis,
  Franzon, Bucher, Basavarajaiah, Oh, and Jenkal}]{FreePDK45}
Stine, J.~E.; Castellanos, I.; Wood, M.; Henson, J.; Love, F.; Davis, W.~R.;
  Franzon, P.~D.; Bucher, M.; Basavarajaiah, S.; Oh, J.; and Jenkal, R. 2007.
\newblock FreePDK: An Open-Source Variation-Aware Design Kit.
\newblock In \emph{2007 IEEE International Conference on Microelectronic
  Systems Education (MSE'07)}, 173--174.

\bibitem[{{Synopsys, Inc.}(2024)}]{TOOL-Proteus-OPC}
{Synopsys, Inc.} 2024.
\newblock Proteus Solutions.
\newblock
  \url{https://www.synopsys.com/manufacturing/mask-solutions/proteus.html}.

\bibitem[{Tsai, Liu, and Ren(2023)}]{tsai2023rtlfixer}
Tsai, Y.; Liu, M.; and Ren, H. 2023.
\newblock {RTLFixer}: Automatically fixing rtl syntax errors with large
  language models.
\newblock \emph{arXiv preprint arXiv:2311.16543}.

\bibitem[{Wu et~al.(2024)Wu, He, Zhang, Yao, Zheng, Zheng, and
  Yu}]{wu2024chateda}
Wu, H.; He, Z.; Zhang, X.; Yao, X.; Zheng, S.; Zheng, H.; and Yu, B. 2024.
\newblock {ChatEDA}: A large language model powered autonomous agent for eda.
\newblock \emph{IEEE Transactions on Computer-Aided Design of Integrated
  Circuits and Systems}.

\bibitem[{Wu et~al.(2016)Wu, Kwa, Wan, Wang, John, Deeth, Chen, Cecil, Meng,
  and Lucas}]{BBS-recipe-Wu-2016}
Wu, L.; Kwa, D.; Wan, J.; Wang, T.; John, M.~S.; Deeth, S.; Chen, X.; Cecil,
  T.; Meng, X.; and Lucas, K. 2016.
\newblock {Building block style recipes for productivity improvement in OPC,
  RET and ILT flows}.
\newblock In \emph{Design-Process-Technology Co-optimization for
  Manufacturability X}, volume 9781. SPIE.

\bibitem[{Yang et~al.(2020)Yang, Li, Deng, Ma, Yu, and
  Young}]{OPC-TCAD2020-Yang}
Yang, H.; Li, S.; Deng, Z.; Ma, Y.; Yu, B.; and Young, E. F.~Y. 2020.
\newblock {GAN-OPC}: Mask Optimization with Lithography-guided Generative
  Adversarial Nets.
\newblock \emph{IEEE Transactions on Computer-Aided Design of Integrated
  Circuits and Systems (TCAD)}.

\bibitem[{Yang and Ren(2024)}]{pmlr-v235-yang24s}
Yang, H.; and Ren, H. 2024.
\newblock {ILILT}: Implicit Learning of Inverse Lithography Technologies.
\newblock In Salakhutdinov, R.; Kolter, Z.; Heller, K.; Weller, A.; Oliver, N.;
  Scarlett, J.; and Berkenkamp, F., eds., \emph{Proceedings of the 41st
  International Conference on Machine Learning}, volume 235 of
  \emph{Proceedings of Machine Learning Research}, 56319--56331. PMLR.

\bibitem[{Zheng et~al.(2023)Zheng, Ma, Zhu, Chen, Zhao, Yin, Yu, and
  Yu}]{OPC:OpenILT}
Zheng, S.; Ma, Y.; Zhu, B.; Chen, G.; Zhao, W.; Yin, S.; Yu, Z.; and Yu, B.
  2023.
\newblock {OpenILT}: An Open-source Platform for Inverse Lithography Technique
  Research.
\newblock \url{https://github.com/OpenOPC/OpenILT/}.

\bibitem[{Zhu et~al.(2023{\natexlab{a}})Zhu, Zheng, Yu, Chen, Ma, Yang, Yu, and
  Wong}]{L2OILT_TCAD}
Zhu, B.; Zheng, S.; Yu, Z.; Chen, G.; Ma, Y.; Yang, F.; Yu, B.; and Wong, M.
  2023{\natexlab{a}}.
\newblock L2O-ILT: Learning to Optimize Inverse Lithography Techniques.
\newblock \emph{IEEE Transactions on Computer-Aided Design of Integrated
  Circuits and Systems}.

\bibitem[{Zhu et~al.(2023{\natexlab{b}})Zhu, Xue, Chen, Zhou, Tang, Schuurmans,
  and Dai}]{zhu2023large}
Zhu, Z.; Xue, Y.; Chen, X.; Zhou, D.; Tang, J.; Schuurmans, D.; and Dai, H.
  2023{\natexlab{b}}.
\newblock Large language models can learn rules.
\newblock \emph{arXiv preprint arXiv:2310.07064}.

\end{thebibliography}
